\documentclass[conference]{IEEEtran}
\usepackage{graphicx}
\usepackage{subfigure}

\ifCLASSINFOpdf
\else
\fi
\begin{document}
%
\title{A Personalized Diagnostic Generation Framework Based on Multi-source Heterogeneous Data}


\author{
\IEEEauthorblockN{Jialun Wu$^{1}$, Zeyu Gao$^{1}$, Haichuan Zhang$^{1,2}$, Ruonan Zhang$^{3}$, Tieliang Gong$^{1}$, Chunbao Wang$^{4}$, and Chen Li$^{1,*}$}
\IEEEauthorblockA{$^{1}$School of Computer Science and Technology, Xi'an Jiaotong University, Xi'an, China\\
National Engineering Lab for Big Data Analytics, Xi'an Jiaotong University, Xi'an, China} 
\IEEEauthorblockA{$^{2}$School of Electrical Engineering and Computer Science, Pennsylvania State University, University Park, PA, USA}
\IEEEauthorblockA{$^{3}$Library of Xi'an Jiaotong University, Xi'an, China}
\IEEEauthorblockA{$^{4}$Department of Pathology, the First Affiliated Hospital of Xi'an Jiaotong University, Xi'an, China\\
Email: andylun96@stu.xjtu.edu.cn}
}


%


\maketitle

\begin{abstract}
Personalized diagnoses have not been possible due to sear amount of data pathologists have to bear during the day-to-day routine. This lead to the current generalized standards that are being continuously updated as new findings are reported. It is noticeable that these effective standards are developed based on a multi-source heterogeneous data, including whole-slide images and pathology and clinical reports. In this study, we propose a framework that combines pathological images and medical reports to generate a personalized diagnosis result for individual patient. We use nuclei-level image feature similarity and content-based deep learning method to search for a personalized group of population with similar pathological characteristics, extract structured prognostic information from descriptive pathology reports of the similar patient population, and assign importance of different prognostic factors to generate a personalized pathological diagnosis result. We use multi-source heterogeneous data from TCGA (The Cancer Genome Atlas) database. The result demonstrate that our framework matches the performance of pathologists in the diagnosis of renal cell carcinoma. This framework is designed to be generic, thus could be applied for other types of cancer. The weights could provide insights to the known prognostic factors and further guide more precise clinical treatment protocols.
\end{abstract}



%
\IEEEpeerreviewmaketitle

\section{Introduction}

So far, the pathological diagnosis of cancer tissue sections has been made based on general and standardized guidelines. 
However, precision medicine would clearly benefit cancer patients since heterogeneity of the disease indicates that patients with the same symptoms may have different underlying illnesses—linked to different genetic mutations. With the increasing incidence of cancer, we explored new ways to guide the diagnosis and prognosis of cancer. In prognoses, survival prediction can be considered essential. Accurately predicting the survival period of cancer patients is crucial to treatment and rehabilitation planning for clinicians and the psychological welfare of patients. There are widespread uses of medical text processing and whole-slide images analysis \cite{1}, and automated computational histopathological analysis systems have shown great promises in diagnosis in recent years. They are being used in various cancers, including kidney cancer \cite{2}, breast cancer \cite{3}, lung cancer \cite{4}, and prostate cancer \cite{5}. The mentioned works are fueled by a large amount of multi-modal heterogeneous data, including histopathological whole-slide images (WSIs), pathology reports and such specialized data sets which are being accumulated over the years. Information extracted from images is close to genomic data \cite{6}. Thus, we will utilize the similarity of the images to find a better group of patients with more genetic similarities. The group could then be used as a basis for prognosis factors weight and survival analysis.

Histopathological images have long been central to cancer diagnosis, staging, and prognosis, pathologists widely use them in clinical practice. In clinics, the pathological diagnostic report is an important part of the overall clinical diagnosis and the clinicians' foundation for providing patients with the best of care. However, the shortage of new and experienced pathologists is common in modern pathology \cite{7}. The diagnosis and treatment of some diseases often lack precise pathological examination results and are only based on speculation with incomplete evidence. It takes a significant amount of time to train a pathologist and even longer for pathologists to become experienced and diagnose different types of tumors.

The pathology report is an essential unstructured clinical document that is mainly composed of textual data recorded by pathologists in natural language. The reports often include necessary information on patients, macroscopic specimens, and microscopic specimens. There are various pathological characteristics in the reports—prognostic factors, for example, cancer type and subtype, anatomic site, tumor size, histological grade, and the TNM stage. The prognostic factors play a crucial role in cancer treatment planning. For different patients, different factors have different influences on diagnosis and prognosis. In this study, we will explore the importance of each prognostic factor by its effects on the survival outcome of a personalized group of patients based on their histopathological similarity.

The mentioned multi-source heterogeneous data can significantly improve the performance of survival predictions. Therefore, methods for effectively combining the multi-modal data to improve cancer survival prediction is needed. In breast cancer, a recent study \cite{8} has developed a multi-state statistical model and used to simulate the different phases of breast cancer, including local recurrence, distant recurrence, breast cancer-related death, and other reasons. The model utilizes the patient’s age, tumor size, and the level of clinical variables affecting the survival probability to predict the individual risk of a recurrence. In pulmonary adenocarcinoma, the TNM stage is the most important predictive factor. 
In the US, the Mayo Clinic uses a scoring system that consisted of tumor stage, size, nuclear grade, and necrosis to classifies clear cell carcinomas into low-risk, medium-risk, and high-risk groups. Though the mentioned works can predict treatment outcomes, none of which leverage a specialized group of patients to arrive at a more precise prediction with a specialized group of patients.
To tackle these problems, we curate a set of pathology images with their corresponding pathology reports and clinical reports from the TCGA (the cancer genome atlas) project \cite{9}. Since there are many cancer types in the TCGA data set, we scope our work to the renal cell carcinoma (RCC), which is one of the most common malignant tumors in adult kidney type \cite{10}. It has three common histologic subtypes: clear cell carcinoma (ccRCC), papillary carcinoma (pRCC), and chromophobe carcinoma (chRCC) \cite{11}. Each case in the TCGA Project has its corresponding pathological report and clinical information. The reports are made after pathologists observed the histopathological sections stained with hematoxylin and eosin (H\&E) under a microscope to make the diagnosis by inspecting the morphological characteristics of histology slides \cite{12}. With a pathology report, clinicians can make informed and precise decisions about treatments with an overall understanding of patients’ symptoms.

In this paper, we demonstrate that a more effective survival analysis of a patient could be made based on a group of histologically similar patients, curated via similar image search algorithms. We develope a predictive model that automatically assigns weights to the prognostic factors of each patient, enabling personalized pathology reports for individual patients. Furthermore, the framework can assess personalized patients' survival risk. We believe that this is the first attempt to co-analyze histopathological features from whole-slide images with prognostic factors from pathology reports to generate personalized weights and survival risks. A new type of pathology reports, personalized, that emerge from our framework could assist pathologists and oncologists to understand the significance of each prognostic factors for each patient. These weighted factors may also be useful for new pathologists’ training. The results show that the predictive performance was similar to that given by pathologists with long working experiences. We believe that our framework could be considered data-independent. Our work could fundamentally alleviate the problem of long training cycles for pathologists, and further research will help demonstrate that the same approach for other types of cancer (such as lung cancer, prostate cancer, and breast cancer) can calculate the weight of his target characteristics and predict his survival risk.

The organization of this paper is as follows: The second part discusses the data set of pathological images and pathological reports and describes our framework and experimental process in detail. In the third part, the results section introduces the statistics of our data set and the performance of our framework, and in the last two part, we discuss the experimental results and limitations.

\begin{figure*}[]
\centering
\includegraphics[width=180mm]{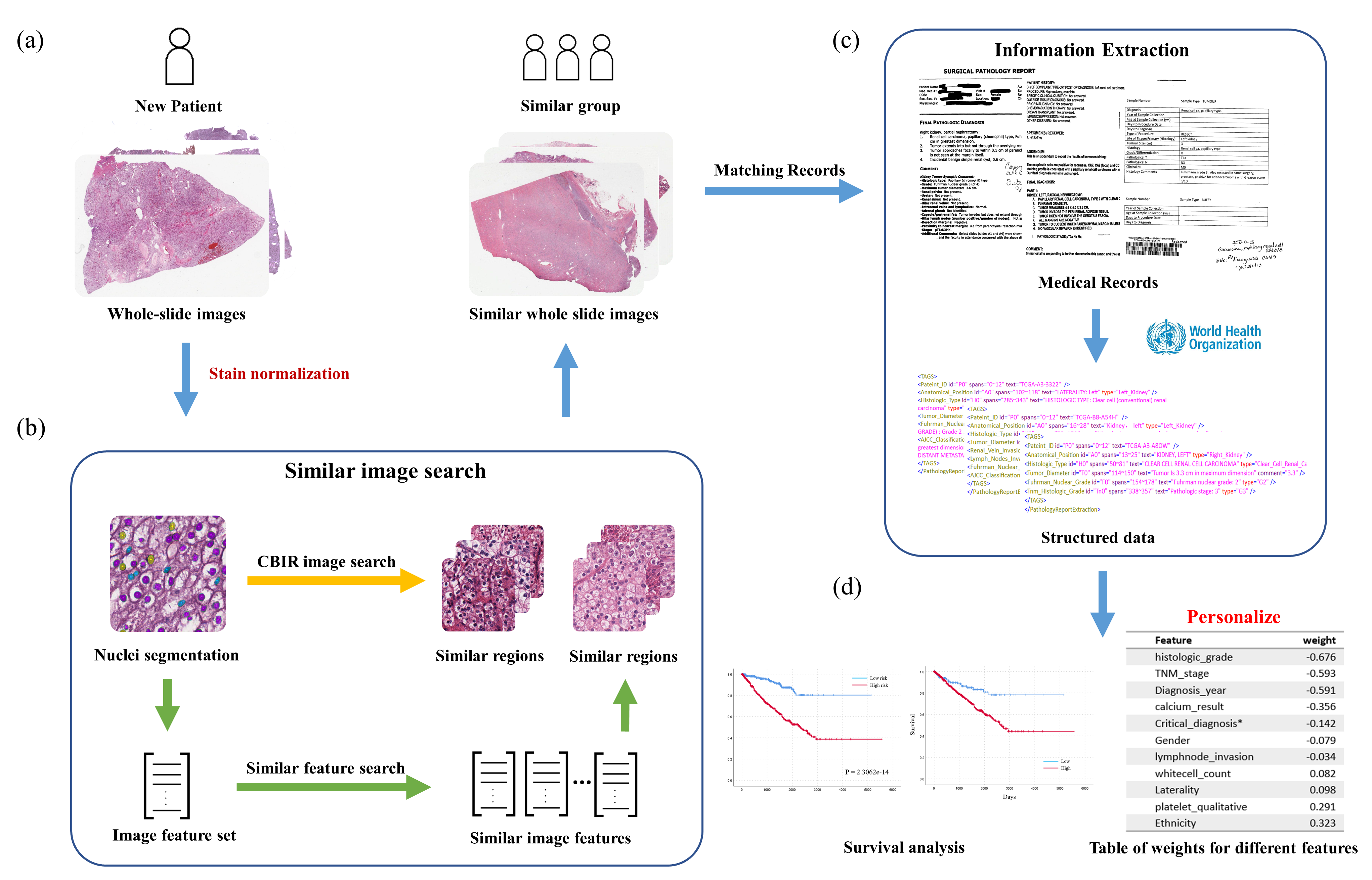}
\caption{The workflow of the framework. (a) In order to search for the simlar population group of the patient, the whole-silde images with stain normalization of the correspongding patient are used as input. (b) Similar image search function: The first method is content-based image retriveal using deep learning, which used patches from whole-slide images as input and the DeepRank network pretrained on ImageNet to get the similar images. The second method is to extract the histological characteristics from the patches and search for the similar image features, which correspond to the similar patches. The two groups of similar patches obtained in two method were integrated and corresponding to WSIs to further obtain similar WSIs information, and similar patient population groups information. (c) Finding pathological reports and clinical information of corresponding patients according to similar patient population and using information extraction techniques to extract structured data from descriptive pathology reports according to the World Health Organization's guideline. (d) Calculating weights for prognostic factors and survival analysis based on prognosis factors.}
\label{f1}
\end{figure*}

\begin{figure*}[]
\centering
\subfigure{
\begin{minipage}[]{\textwidth}
\centering
\includegraphics[width=0.1\textwidth]{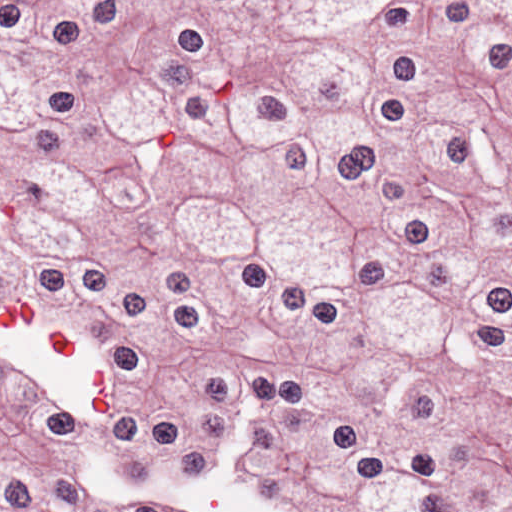}
\includegraphics[width=0.1\textwidth]{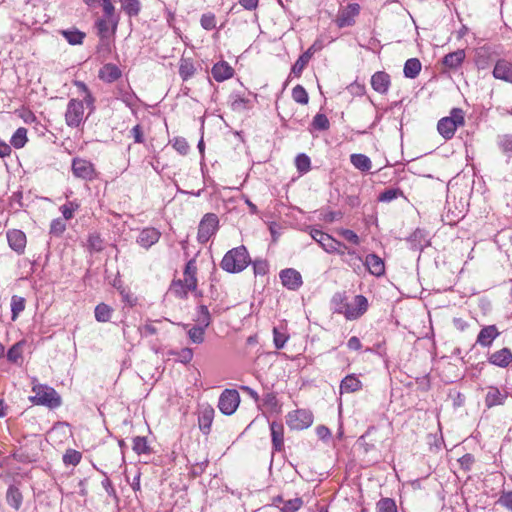}
\includegraphics[width=0.1\textwidth]{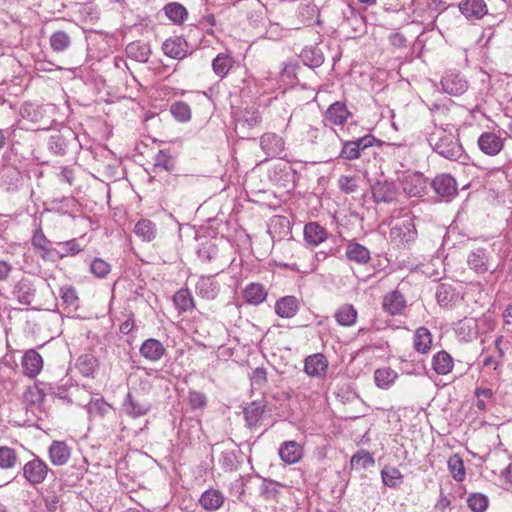}
\includegraphics[width=0.1\textwidth]{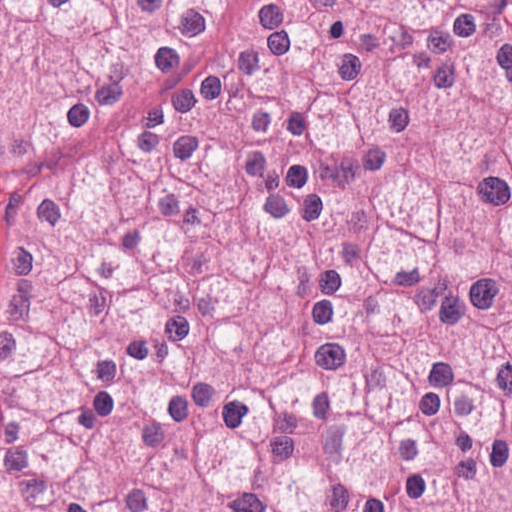}
\includegraphics[width=0.1\textwidth]{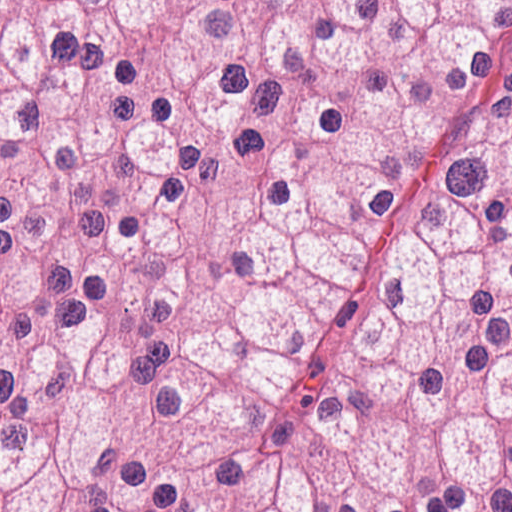}
\includegraphics[width=0.1\textwidth]{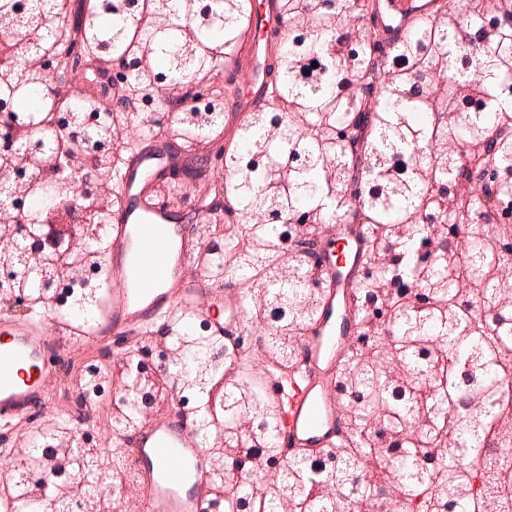}
\end{minipage}
}
\subfigure{
\begin{minipage}[]{\textwidth}
\centering
\includegraphics[width=0.1\textwidth]{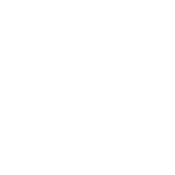}
\includegraphics[width=0.1\textwidth]{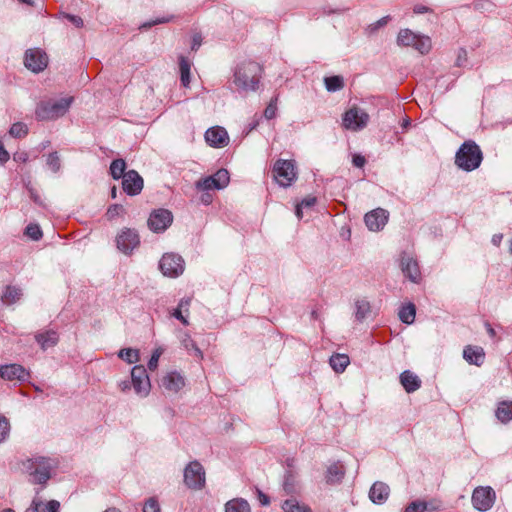}
\includegraphics[width=0.1\textwidth]{image/fig222.png}
\includegraphics[width=0.1\textwidth]{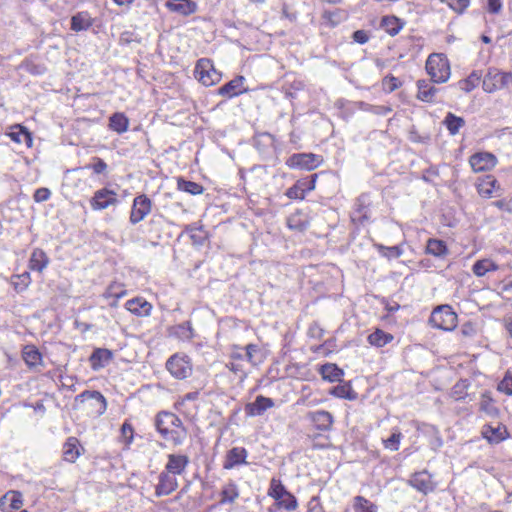}
\includegraphics[width=0.1\textwidth]{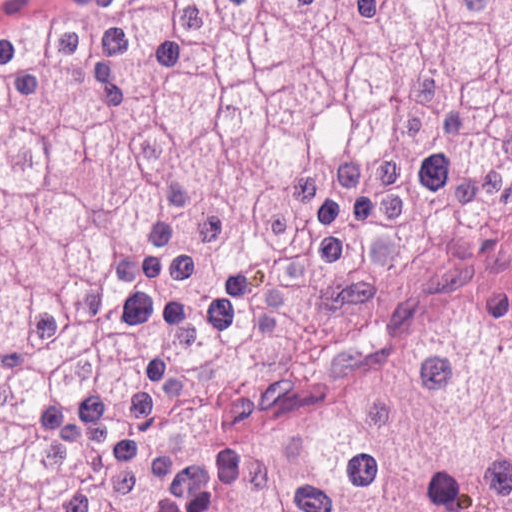}
\includegraphics[width=0.1\textwidth]{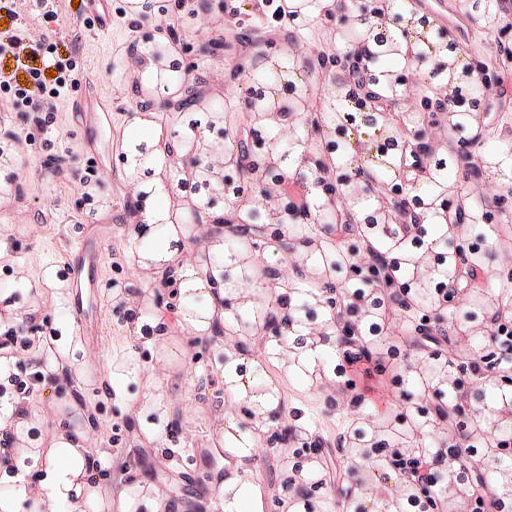}
\end{minipage}
}
\subfigure{
\begin{minipage}[]{\textwidth}
\centering
\includegraphics[width=0.1\textwidth]{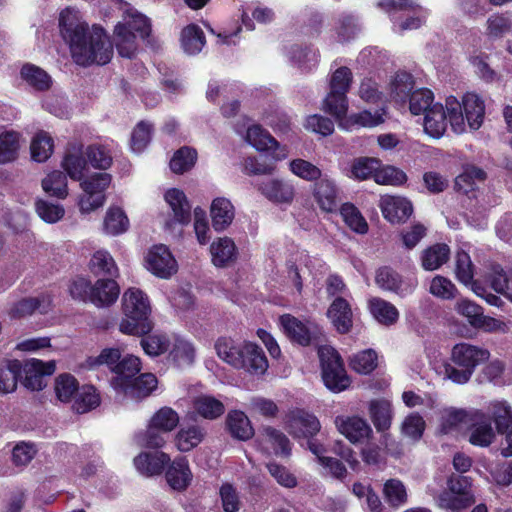}
\includegraphics[width=0.1\textwidth]{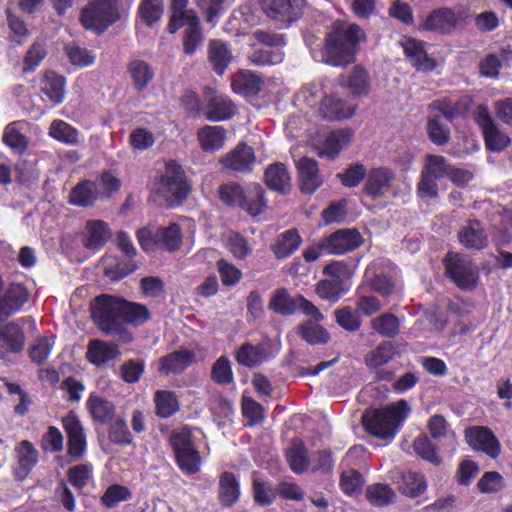}
\includegraphics[width=0.1\textwidth]{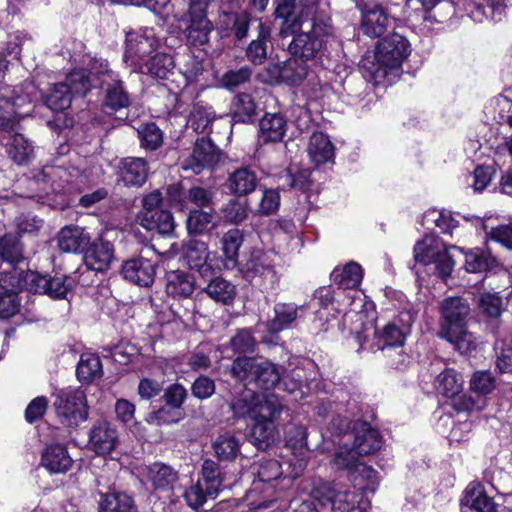}
\includegraphics[width=0.1\textwidth]{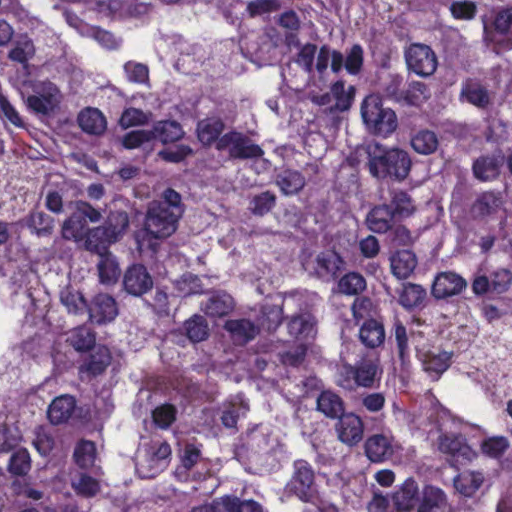}
\includegraphics[width=0.1\textwidth]{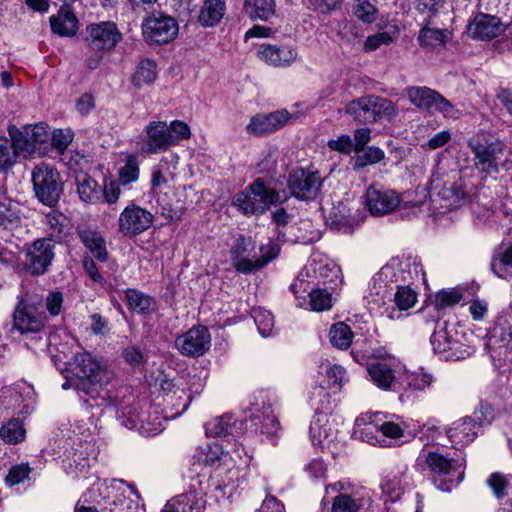}
\includegraphics[width=0.1\textwidth]{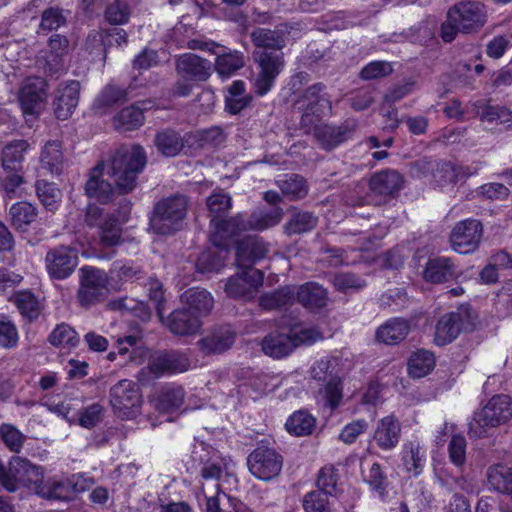}
\end{minipage}
}
\subfigure{
\begin{minipage}[]{\textwidth}
\centering
\includegraphics[width=0.1\textwidth]{image/blank.png}
\includegraphics[width=0.1\textwidth]{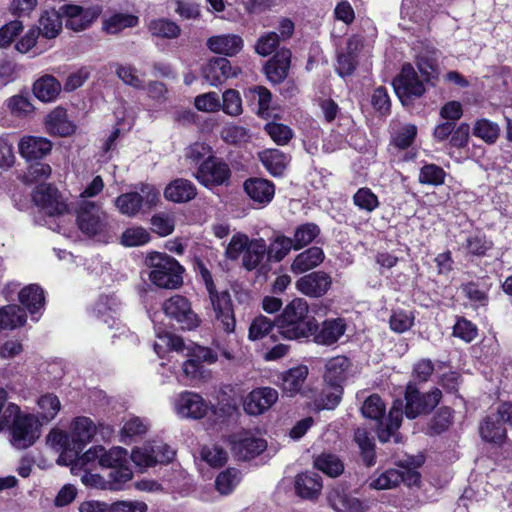}
\includegraphics[width=0.1\textwidth]{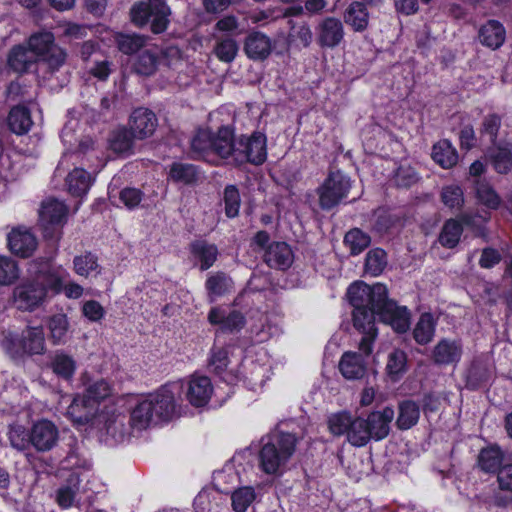}
\includegraphics[width=0.1\textwidth]{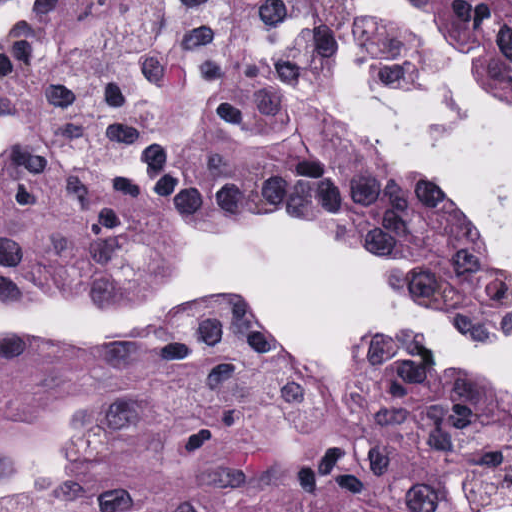}
\includegraphics[width=0.1\textwidth]{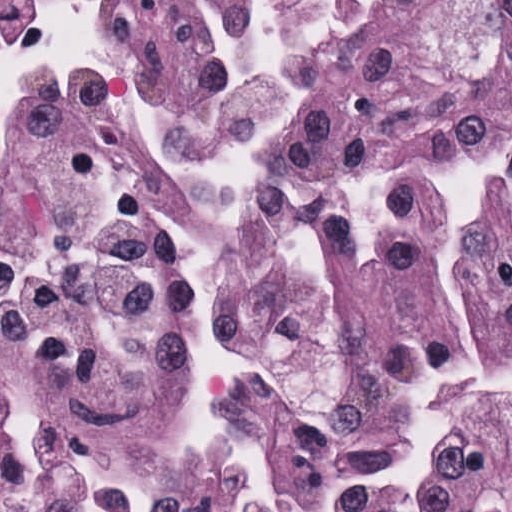}
\includegraphics[width=0.1\textwidth]{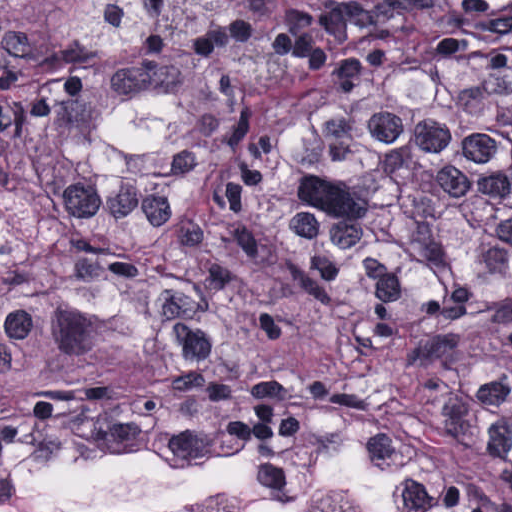}
\end{minipage}
}
\caption{The top five similar images retrieval results of two different regions. 
Method: (a) Deep Ranking network with deep learning method, (b)  Similar image search based on image features.}
\label{f2}
\end{figure*}

\section{Materials and Methods}

In order to calculate the weight of different prognostic factors for each patient, a group of population with similar histology is needed for each case. We explore deep learning and computing image feature similarity two options for establishing the groups. For both methods, the top 500 images similar to the ground image are used as a proxy to obtain corresponding pathological reports and clinical information. For the cases with valid clinical information, the main methods we use for survival analysis of cancer patients include Kaplan-Meier survival analysis curve \cite{22}, Cox proportional risk model \cite{23} to calculate the weights of each clinical feature. It is possible to find similar tumor patient groups for each patient, and assign weights to the pathological and clinical characteristics of the patients to more accurately customize the weight of prognostic factors.

\subsection{Data sources and annotations}
Based on TCGA project, we select kidney cancer from more than twenty types of cancer in our study. TCGA collects histopathological images, detailed pathological reports, and clinical information of patients with various cancer types to jointly display tumor characteristics and results of patients. In this paper, we collect histopathology image data, pathological diagnosis report data, and clinical data from TCGA for kidney cancer, including clear cell renal cell carcinoma, papillary renal cell carcinoma, and chromophobe renal cell carcinoma. Specifically, we selected 3281 digital whole-slide images from 943 patients and 894 pathology reports in renal cancer. By matching images of the same patients with reports, we ended up with pathology reports of 859 patients in renal cancer that could be matched with digital pathology slides (two or more per person).

\subsection{The proposed framework}
Fig. 1 depicts our data analysis workflow, with the following sections describing the information for each step. OpenHI \cite{32} is an open-source annotation and processing software which specially designed for histopathology images. The pathologists can perform routine operations such as area selection and annotation on this software. Now, it supports several types of annotation and can record the time of annotation. At the beginning of the introduction to our framework, the normalization of haematoxylin and eosin stain was made at the training and validation stages using a reference image and the Macenko algorithm \cite{33}.

\subsection{Similar histopathological image search}
Medical image retrieval technology is a process in which the underlying physical features of the image to be retrieved (such as gray scale, texture, shape, spatial position relation) are extracted, and then compared with the feature vectors in the database according to the similarity measurement method, and the preceding most similar image is returned to the user in the order of decreasing similarity. In our similar pathological image search, we use two methods: similar image search based on computing image feature similarity and similar image search based on deep learning.

In the first method, we use the method based on Deep Ranking \cite{35}. In the process of similar image model training, we use three different images to be transferred to the model. The first two images belong to the same category, and the third image belongs to different categories. The model calculates the embedding distance between each image pair in the three images and makes a comparison. Through different categories of natural images in ImageNet \cite{36}, the model we use learns to distinguish between similar and dissimilar images by calculating and comparing the embedding distance of the input images. In the query matching in our pathological image data set, the model first calculates the embedding of the selected input image, and then calculates its embedding value with other pathological images stored in the data set, and then selects the first 500 images with the highest embedding distance as similarity pathological image.

In the second method, we use interpretable image features to search for the similar images. Our pathological image feature extraction process includes two stages: nuclear segmentation and cellular features extraction. Usually, pathologists will select 8-10 regions of approximately 512-by-512 pixel (40x magnification) in a whole-slide image, and a single patient will be diagnosed with multiple slides (normally 2-4 slides). Histopathological image consisted of tumorous and normal regions. The selected sample region for the search did not only contains tumor cells but also includes inflammatory cells in some image areas in addition to tumor cells. Then we applied the supervised segmentation algorithm (Hovernet) \cite{37} to segment the nucleus from the image. Subsequently, we extracted 31 histological image characteristics for each divided nucleus according to the guideline for characteristics extraction \cite{38}. There contains 10 morphological characteristics, 5 intension-based characteristics, and 16 texture-based characteristics, morphological characteristics describe the shape and size changes of the nucleus. Intension-based characteristics (first-order statistical features) describe the distribution of color changes in the nucleus—mean, median, standard deviation, skewness, and kurtosis; texture-based characteristics (second-order statistical features) quantitatively describe patterns and texture of pixel values: co-occurrence based features and run-length based features. Co-occurrence based features include correlation, cluster shade, cluster prominence, energy, entropy, Haralick correlation, inertia, and inverse difference moment \cite{39}. Run-length based features include gray-level non-uniformity, run-length non-uniformity, low and high gray-level run emphasis, short-run low and high gray-level emphasis, and long-run low and high gray-level emphasis \cite{40}.

After we get the 31 different characteristics of each image. We compare the feature of the search image with the feature of each patch in the data set to calculate their distance. We adopt a method combining dynamic weight adjustment with weighted Markov distance, modify the retrieval strategy, improve the precision of retrieval, overcome the problem of attribute correlation between feature vectors in Euclid distance calculation, and improve the retrieval accuracy. We adopt the method of multi-texture feature fusion to treat the feature vectors in the feature space differently according to different weights.The weight can better reflect the importance of different feature vectors to target retrieval. In the process of constant optimization of query vectors, the previous query vectors are regarded as positive examples and added to the current feedback positive examples to form a more complete set. Meanwhile, the weight is dynamically adjusted until it becomes stable and optimized. In order to improve the computational speed and reduce the computational complexity, we use the FASII similarity search library proposed by Facebook AI Lab \cite{41}, It includes the algorithm of researching in any size vector set, which can improve the speed of searching similar pathological images according to the pathological feature matrix. After obtaining similar images by these two methods, we find which whole-slide image these patches belong to and find the corresponding medical information of these patches in turn according to the corresponding relationship between images, pathological reports, and clinical reports, and convert them into structured text data. Each feature and attribute value corresponding to each feature in the relevant pathology report and clinical information can be obtained from structured textual data.

\subsection{Histopathological prognostic factors extraction from pathology reports}
The selected TCGA pathology reports included 894 cases of renal cell carcinoma are carefully annotated by pathologists. The content of annotations involve several attributes: cancer type, tumor resection site preference, cancer subtypes, maximum tumor diameter, histological grade, TNM stage, Description of lymph node metastasis is essential in all cancers. For renal cancer images and pathology report, we need to focus on whether the cancer areas invade the renal veins; for lung cancer images and pathology reports, the key concerns are whether tumor cells invade the pulmonary membrane and whether the tumor spreads in the alveolar cavity. Based on two examples of biomedical histopathology corpus annotations \cite{42,43}, we designed an iterative annotation workflow and revised our annotation guideline several times. All annotations were done at the document level so that the annotators can leverage the context in difficult cases. An open-source software called MAE version 2.2.10 was used as the annotation tool throughout the entire process \cite{44}. We sorted out the annotation content of the pathology report and matched them with the pathology pictures based on the same patient ID. We use annotated pathology report data as the training set to learn unannotated pathology reports using an end-to-end two-layer LSTM neural network based on tagging strategy. We use the method we proposed \cite{wu2020structured} to extract the essential information, when we converted unstructured pathology reports and clinical information into structured data, due to incomplete pathology data, we introduced the guidelines of the WHO Guidelines \cite{45} to supplement and improve our structured data, and support related structures in the standards text description information is added to the clinical data.

\subsection{Statistical methods for prognosis prediction}
We establish a univariate Kaplan-Meier curve model \cite{46} for survival analysis and a multivariate Proportional hazards model \cite{47}, according to the last section of the similar image corresponding to the prognostic factors of the query image and the characteristic value to calculate each patient's risk index and give weight to the characteristics of each patient, according to the different characteristics of predict survival and prognosis of the patients. In our experiment, the more accurate the prognosis prediction of the attribute, the higher the corresponding weight of the attribute.

Regression Models in our Regression risk model, We assume that $T$ is time, $h(t,X)$ is the risk rate when the current time is $T$, and the objective factor is $X$, then:
\begin{equation}
    h(t,X)=\lambda_{0}(t) exp (\beta \cdot X)
\end{equation}
By logarithmic operation of the formula, we can get:
\begin{equation}
    ln(h(t,X))=\beta \cdot X + ln(\lambda_{0}(t))
\end{equation}
Suppose there are a total of $N$ events, the risk characteristic of the $ith$ event is $X^{(i)}$ , and the occurrence time is $t_{i}$, from which we get the maximum likelihood function as follows:
\begin{equation}
    L(\beta)=\prod_{i=1}^{N}\frac{exp(\beta \cdot X^{(i)}}{\sum_{j:t_j\geq t_i}^{}exp(\beta \cdot X^{(j)}}
\end{equation}
The logarithmic likelihood function is:
\begin{equation}
    l(\beta)=lnL(\beta)=\sum_{i=1}^{N}[ln\beta \cdot X^{(i)}- ln(\sum_{j:t_j\geq t_i}^{}exp(\beta \cdot X^{(j)})]
\end{equation}
The gradient is:
\begin{equation}
    \frac{\partial l(\beta)}{\partial \beta}=\sum_{i=1}^{N}[X^{(i)}-\frac{\sum_{j:t_j\geq t_i}^{} X^{(j)} \cdot exp(\beta \cdot X^{(j)}}{\sum_{j:t_j\geq t_i}^{}exp(\beta \cdot X^{(j)}}]
\end{equation}
Then we can use the gradient descent method to estimate our parameters.

\begin{table*}[]
\caption{Survival-Associated Prognostic Factors, Identified by The Kaplan-Meier Estimator and Log-Rank Test (P$<$0.05)}
\begin{center}
\begin{tabular}{|l|l|l|l|l|l|l|l|l|}
\hline
Feature                  & P-Value    & P/N & Feature               & P-Value    & P/N & Feature               & P-Value    & P/N \\ \hline
histological\_type       & 2.4023e-7  & P   & ethnicity             & 0.007210   & P   & pathologic            & 8.035e-43  & P   \\ \hline
gender                   & 0.563      & N   & hemoglobin\_result    & 0.000793   & P   & TNM\_stage            & 1.5522e-47 & P   \\ \hline
neoadjuvant\_treatment   & 0.015391   & P   & Critical\_diagnosis   & 0.038198   & P   & tumor\_type           & 0.039763   & P   \\ \hline
laterality               & 0.005815   & P   & histologic\_grade     & 3.7118e-33 & P   & white\_cell\_count    & 0.028612   & P   \\ \hline
neoplasm\_cancer\_status & 3.7698e-47 & P   & platelet\_qualitative & 6.6477e-15 & P   & tissue\_prospective   & 0.142383   & N   \\ \hline
lymphnode\_invasion      & 0.017860   & P   & serum\_calcium        & 0.000309   & P   & tissue\_retrospective & 0.119677   & N   \\ \hline
\end{tabular}
\end{center}
\end{table*}

\begin{figure*}[]
\centering
\subfigure[]{
\centering
\includegraphics[width=0.31\textwidth]{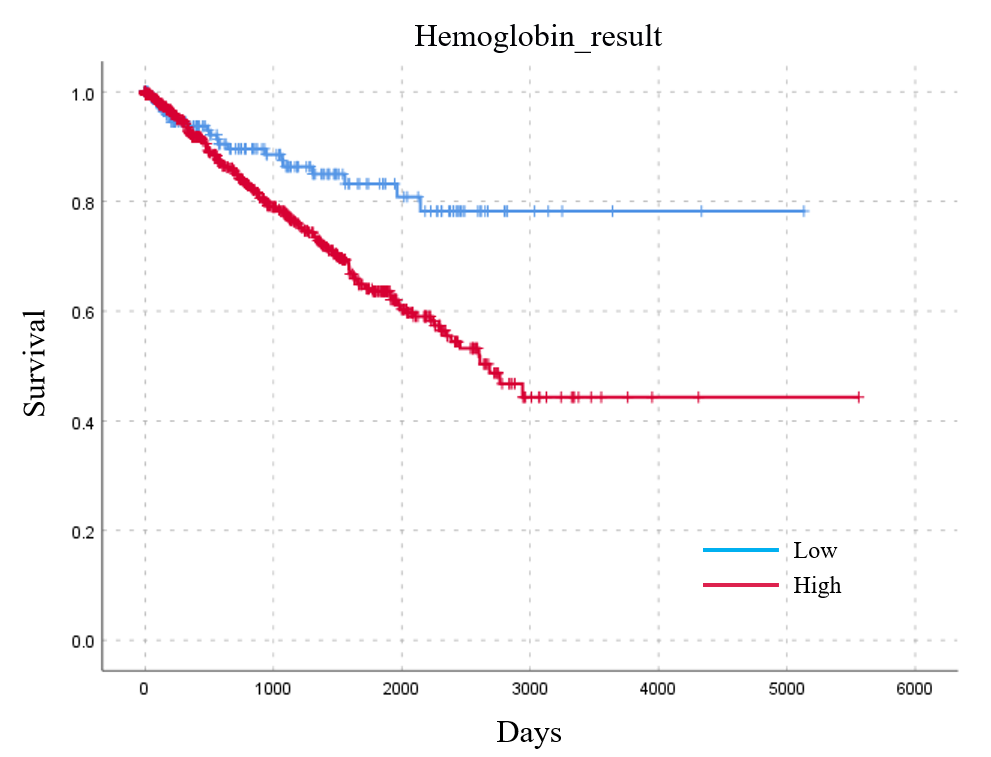}
}
\subfigure[]{
\centering
\includegraphics[width=0.31\textwidth]{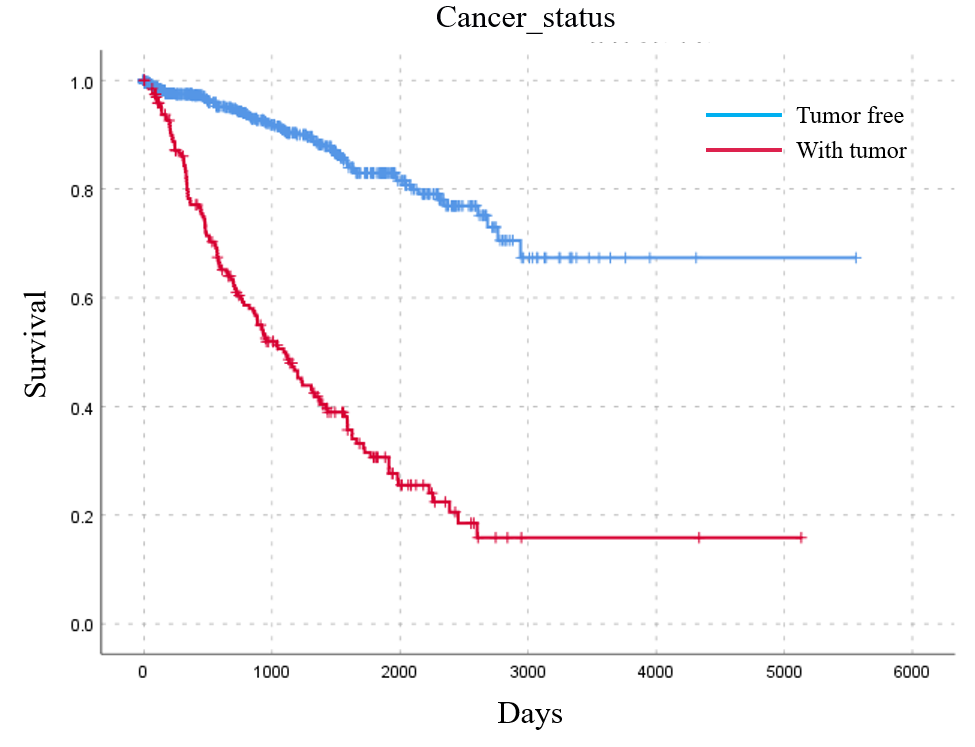}
}
\subfigure[]{
\centering
\includegraphics[width=0.31\textwidth]{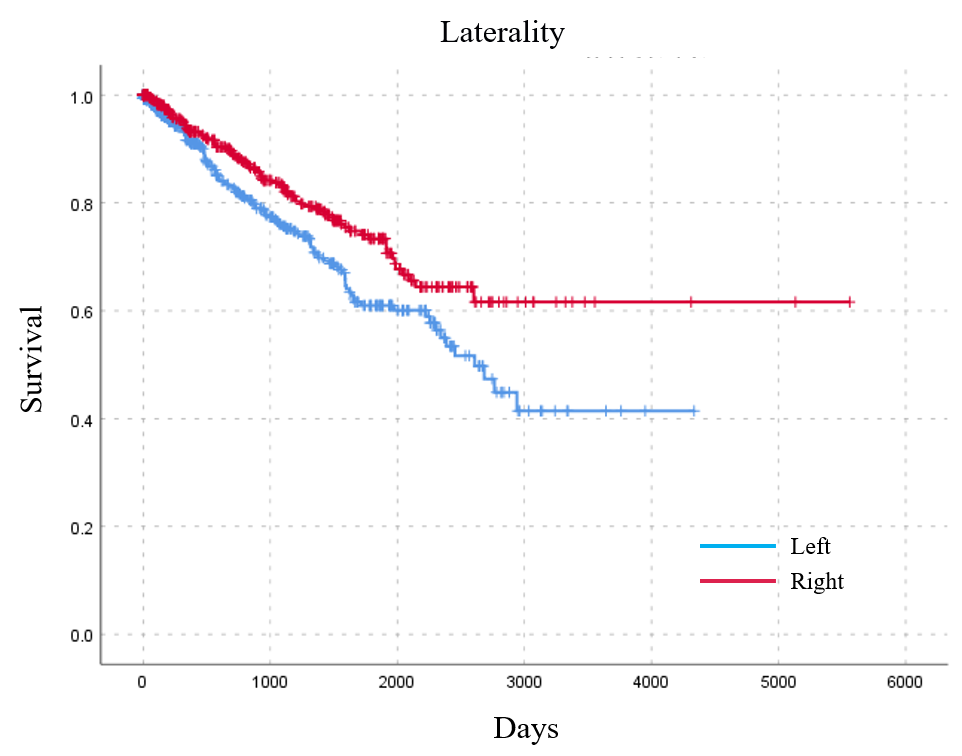}
}
\subfigure[]{
\centering
\includegraphics[width=0.31\textwidth]{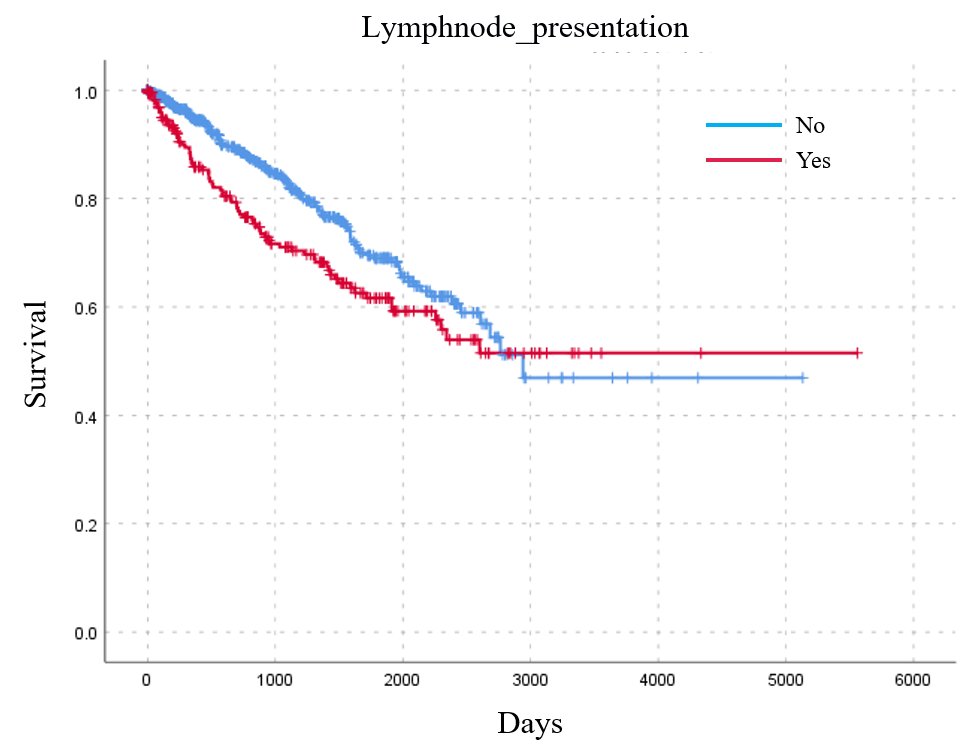}
}
\subfigure[]{
\centering
\includegraphics[width=0.31\textwidth]{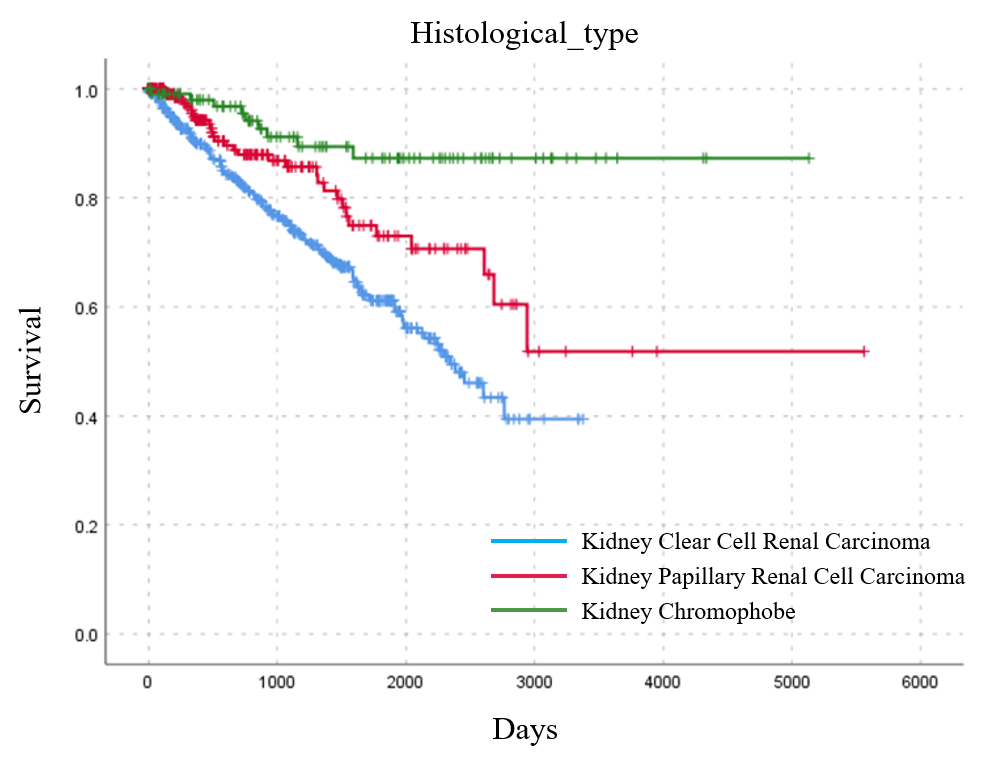}
}
\subfigure[]{
\centering
\includegraphics[width=0.31\textwidth]{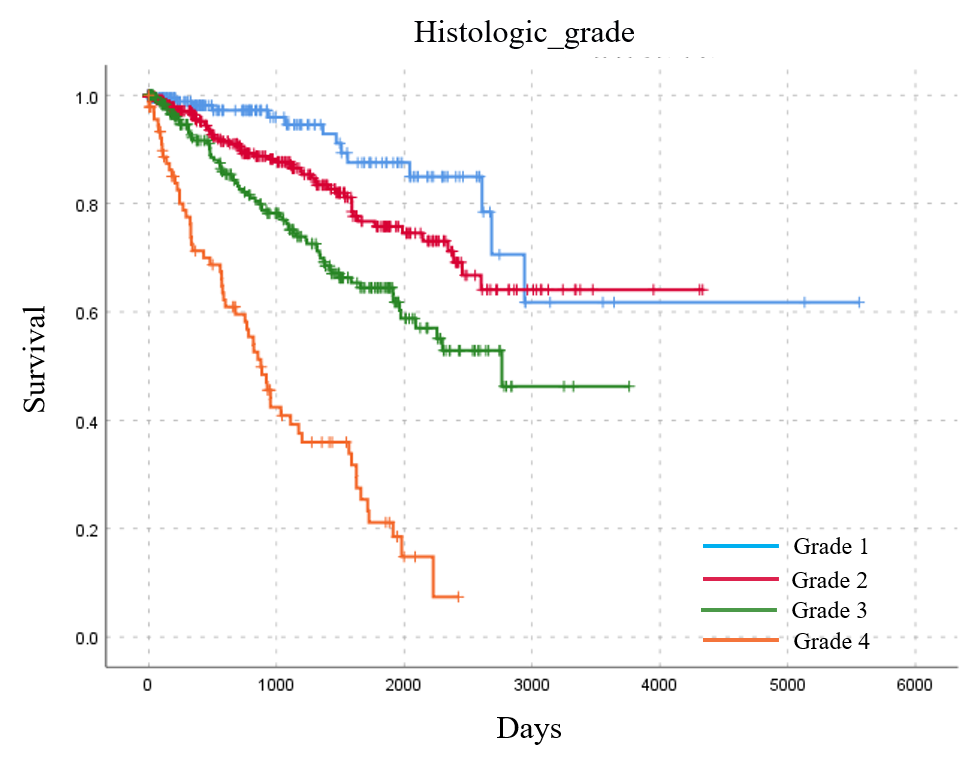}
}

\caption{Prognostic factors predict the survival outcomes of ccRCC patients. Prognostic factors (a) and (b) are variables that have good predictive results. Prognostic factors (c) and (d) can poorly identify prognostic survival results. We still predict survival for multi-factor variables (e) and (f)}
\label{f3}
\end{figure*}

\begin{table*}[]
\caption{Univariate and Multivariate Cox Proportional Hazards Analysis of the Prognostic Values of Lasso-cox Risk Index and Other Prognostic Factors. Hr, Hazard Ratio. Ci, Confidence Interval.}
\begin{center}
\begin{tabular}{|l|l|l|l|l|}
\hline
                                 & \multicolumn{2}{c|}{Univariate Cox regression} & \multicolumn{2}{c|}{Multivariate Cox regression} \\ \hline
Variable                         & \multicolumn{1}{c|}{HR (95\% CI)} & P value    & \multicolumn{1}{c|}{HR (95\% CI)}  & P value     \\ \hline
Lasso-Cox                        & 1.320(0.937-2.308)                & 0.001153   & 1.424(1.262-1.761)                 & 0.018915    \\ \hline
TNM\_stage                       & 1.433(1.219-1.684)                & 0.000013   & 2.096(1.847-2.378)                 & 2.0432e-30  \\ \hline
neoplasm\_histologic\_grade      & 1.287(1.045-1.586)                & 0.017610   & 2.324(1.966-2.747)                 & 4.8673e-23  \\ \hline
person\_neoplasm\_cancer\_status & 3.378(2.380-4.796)                & 9.7962e-12 & 6.413(4.799-8.571)                 & 3.5311e-36  \\ \hline
histological\_type               & 0.551(0.390-0.778)                & 0.000713   & 0.497(0.384-0.644)                 & 1.1895e-7   \\ \hline
serum\_calcium\_result           & 0.767(0.583-1.009)                & 0.005799   & 1.106(0.768-1.344)                 & 0.0010848   \\ \hline
lymphnode\_invasion              & 0.538(0.325-0.891)                & 0.015986   & 1.490(1.019-2.177)                 & 0.039508    \\ \hline
laterality                       & 0.697(0.521-0.931)                & 0.014553   & 0.673(0.507-0.893)                 & 0.006150    \\ \hline
ethnicity                        & 0.241(0.089-0.654)                & 0.005216   & 0.281(0.104-0.756)                 & 0.011934    \\ \hline
gender                           & 0.706(0.524-0.950)                & 0.021716   & 0.918(0.686-1.228)                 & 0.056347    \\ \hline
\end{tabular}
\end{center}
\end{table*}

\begin{figure*}[]
\centering
\subfigure[]{
\centering
\includegraphics[width=0.31\textwidth]{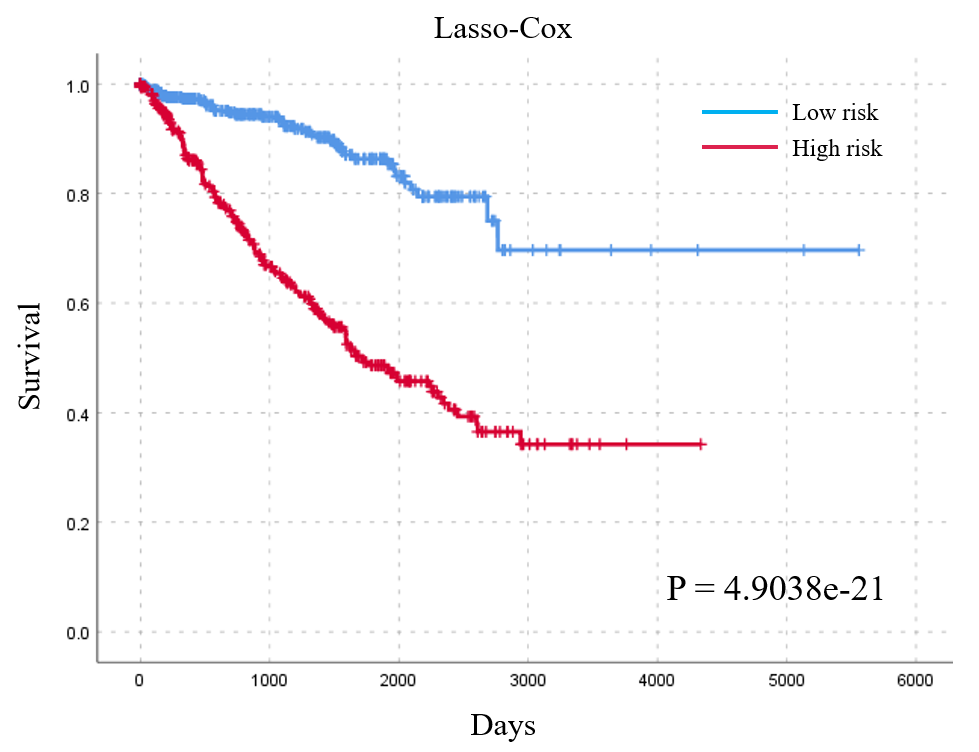}
}
\subfigure[]{
\centering
\includegraphics[width=0.31\textwidth]{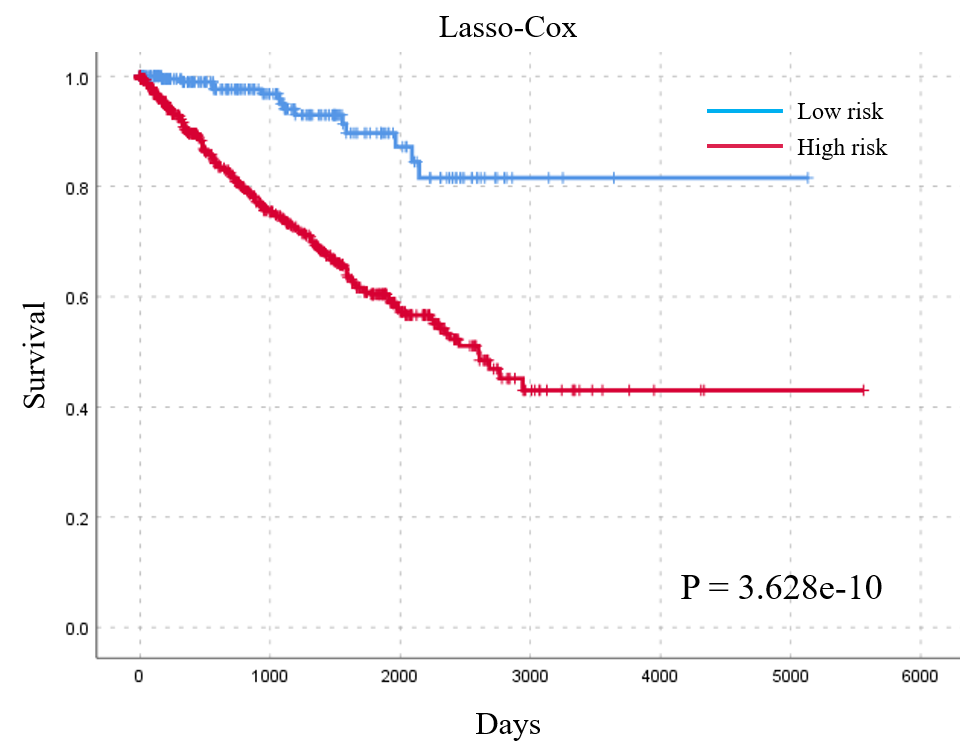}
}
\subfigure[]{
\centering
\includegraphics[width=0.31\textwidth]{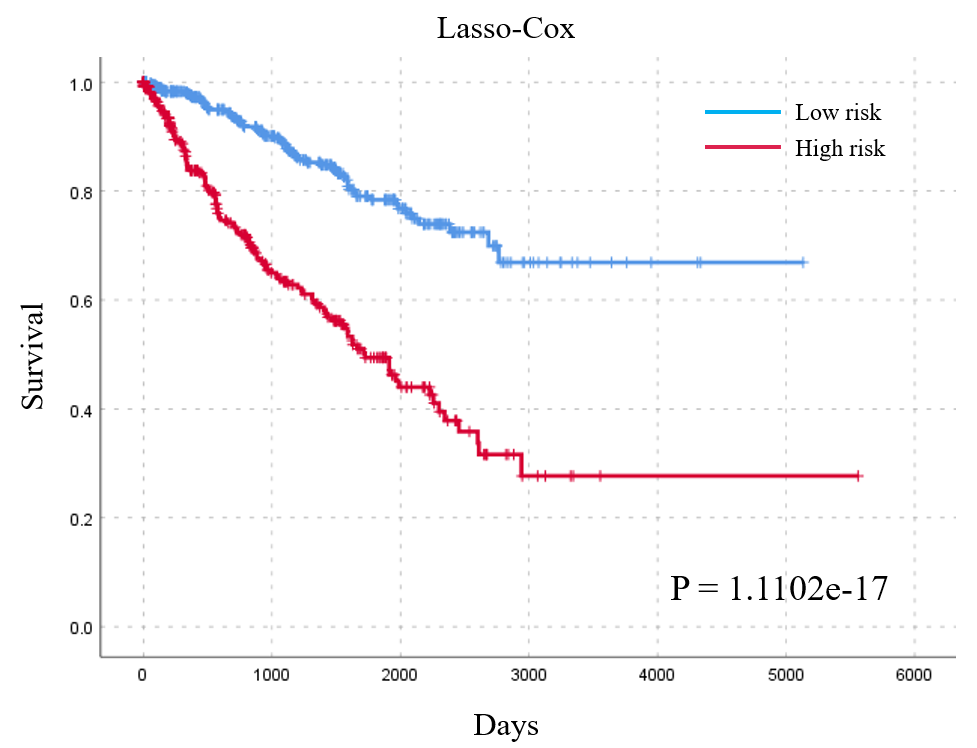}
}

\caption{Integrative analysis of prognosis factors from histopathological images using lasso-Cox can significantly improve the prognosis prediction}
\label{f4}
\end{figure*}

\begin{table*}[]
\caption{Personalized Prognostic Factors Weights for Each Patient}
\begin{center}
\begin{tabular}{|l|r|l|l|r|l|l|r|l|l|r|}
\cline{1-2} \cline{4-5} \cline{7-8} \cline{10-11}
histologic\_grade     & -0.676 &  & Diagnosis\_year     & -0.608 &  & TNM\_stage            & -0.556 &  & Diagnosis\_year       & -0.581 \\ \cline{1-2} \cline{4-5} \cline{7-8} \cline{10-11} 
TNM\_stage            & -0.593 &  & TNM\_stage          & -0.589 &  & Diagnosis\_year       & -0.544 &  & TNM\_stage            & -0.542 \\ \cline{1-2} \cline{4-5} \cline{7-8} \cline{10-11} 
Diagnosis\_year       & -0.591 &  & histologic\_grade   & -0.561 &  & histologic\_grade     & -0.525 &  & histologic\_grade     & -0.488 \\ \cline{1-2} \cline{4-5} \cline{7-8} \cline{10-11} 
calcium\_result       & -0.356 &  & Tumor\_type         & -0.294 &  & calcium\_result       & -0.352 &  & Critical\_diagnosis*  & -0.345 \\ \cline{1-2} \cline{4-5} \cline{7-8} \cline{10-11} 
Critical\_diagnosis*  & -0.142 &  & calcium\_result     & -0.240 &  & lymphnode\_invasion   & -0.352 &  & Tumor\_type           & -0.243 \\ \cline{1-2} \cline{4-5} \cline{7-8} \cline{10-11} 
Gender                & -0.079 &  & lymphnode\_invasion & -0.091 &  & Critical\_diagnosis*  & -0.037 &  & lymphnode\_invasion   & -0.207 \\ \cline{1-2} \cline{4-5} \cline{7-8} \cline{10-11} 
lymphnode\_invasion   & -0.034 &  & Laterality          & 0.020  &  & whitecell\_count      & 0.172  &  & calcium\_result       & -0.175 \\ \cline{1-2} \cline{4-5} \cline{7-8} \cline{10-11} 
whitecell\_count      & 0.082  &  & Gender              & 0.087  &  & Laterality            & 0.205  &  & Gender                & -0.162 \\ \cline{1-2} \cline{4-5} \cline{7-8} \cline{10-11} 
Laterality            & 0.098  &  & whitecell\_count    & 0.097  &  & hemoglobin\_result    & 0.267  &  & whitecell\_count      & 0.310  \\ \cline{1-2} \cline{4-5} \cline{7-8} \cline{10-11} 
platelet\_qualitative & 0.291  &  & Ethnicity           & 0.240  &  & histologic\_type      & 0.303  &  & platelet\_qualitative & 0.325  \\ \cline{1-2} \cline{4-5} \cline{7-8} \cline{10-11} 
Ethnicity             & 0.323  &  & histologic\_type    & 0.336  &  & platelet\_qualitative & 0.405  &  & Ethnicity             & 0.348  \\ \cline{1-2} \cline{4-5} \cline{7-8} \cline{10-11} 
\end{tabular}
\end{center}
\end{table*}

\begin{figure*}[]
\centering
\subfigure[ccRCC grade2]{
\centering
\includegraphics[width=0.22\textwidth]{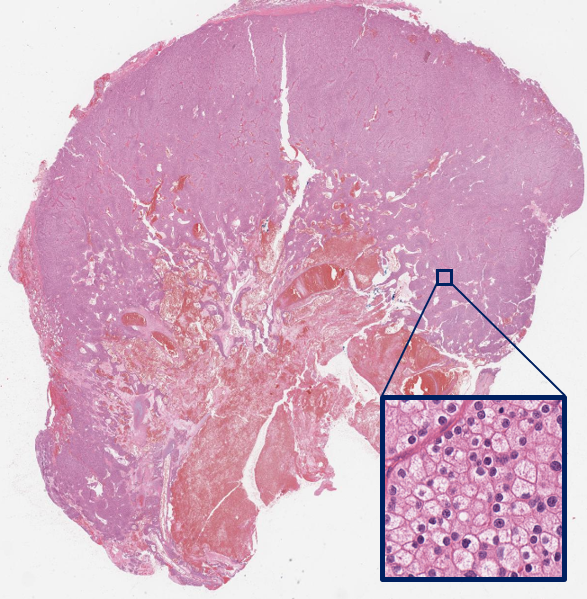}
}
\subfigure[pRCC type2]{
\centering
\includegraphics[width=0.22\textwidth]{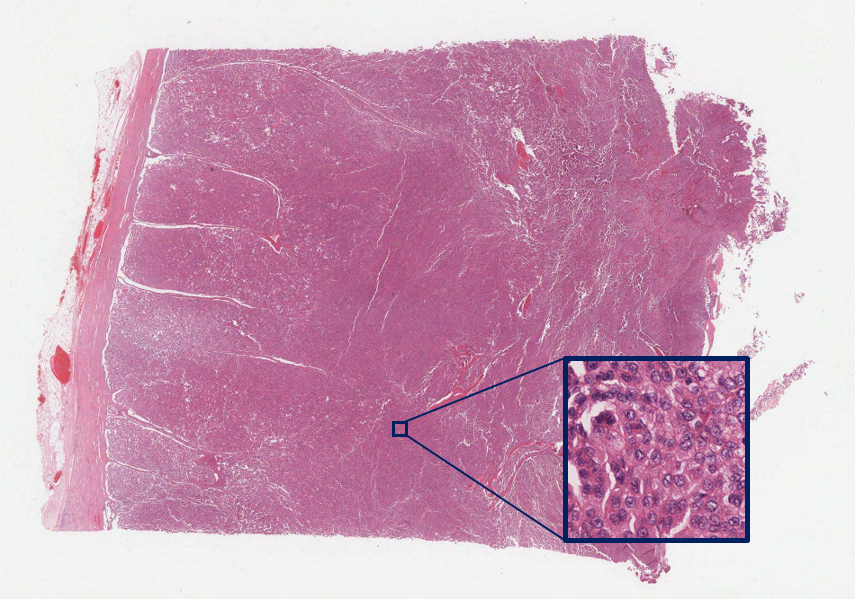}
}
\subfigure[chRCC]{
\centering
\includegraphics[width=0.22\textwidth]{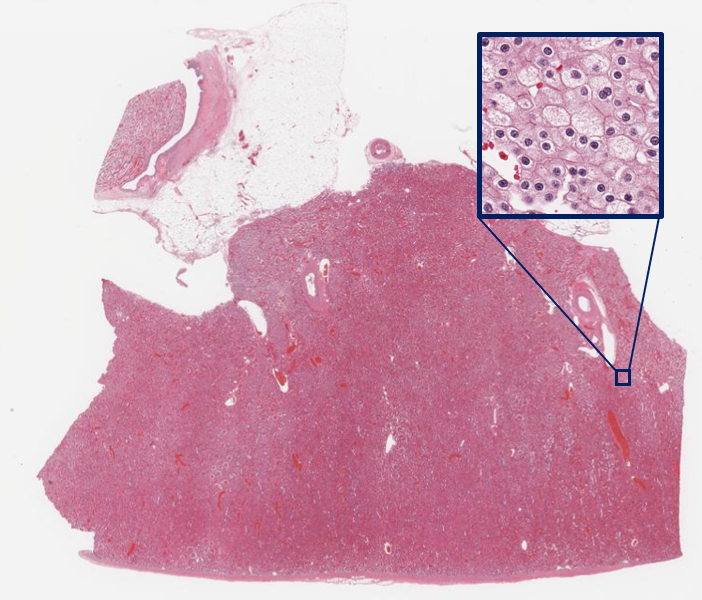}
}
\subfigure[pRCC type1]{
\centering
\includegraphics[width=0.22\textwidth]{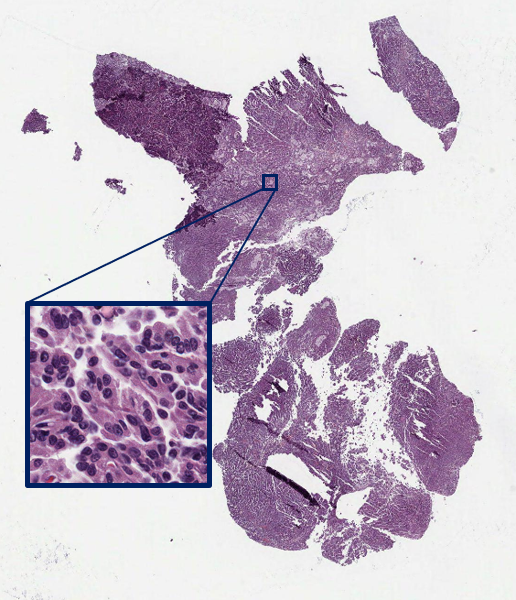}
}

\caption{Patients typical pathology images with select patches correlated to the prognosis factors weights above.}
\label{f5}
\end{figure*}

\section{Result}

\subsection{Similar image retrieval}
In order to find similar population groups of cancer patients, we use deep learning method and image feature-based distance method to retrieve similar images. Taking into account differences in the cancer population, we conduct experiments on two separate data sets. The example retrieval results of the two methods are shown in Fig. 2.

We select the top five most similar pictures from the results as an example and first made a coarse-grained judgment on the similar pictures obtained, respectively, from the subtype and cancer grade. From the example of the results we give, the first image patch is clear cell carcinoma with grade 2. There is almost no other subtypes and non-cancer regions in the corresponding two results, and similar images that rank in the top rarely appear with grade 1 and grade 3 cancer cells. The second image is of type 2 papillary renal cell carcinoma. We search the subtype and corresponding type of each result and found that the subtype rarely has detection errors because the number of images corresponding to papillary is limited. There are also some type1 papillary tumor tissue areas. Then, the pathologists retrieve their corresponding medical information such as age, medical history, to determine whether these similar images can be classified as similar tumor populations. These two methods have precise control over the cell morphology, staining degree and cell distribution of our images, indicating that we have relatively better results in image retrieval of similar tumor patches.

\subsection{Analysis of a single prognostic factor}
In order to study which individual specific prognostic factor is closely related to the survival and prognosis of patients, we first sort each feature according to the size of the feature value, and distinguished the high-risk group from the low-risk group by the median of the feature value. We test the statistical significance of overall survival differences between high-risk groups and low-risk groups. The log-rank test showed that 25 prognostic factors were significantly related to the prognosis (p<0.05). For each prognostic factor, patients are divided into different groups with the median as the dividing point. For P/N, P indicates a positive correlation with survival (i.e., patients with higher characteristic values have a good prognosis), and N indicates a negative correlation with survival. The log test results of all survival-related variables are listed in Table 1. The Kaplan-Meier survival curves of some variables are shown in Figure 3. We select the similar cancer groups of each patient through similar pathological images. Based on similar groups' information, we predict the survival rate of patients in turn for each prognostic factor closely related to the prognosis of patient.

\subsection{Integrating prognostic factors analysis to predict prognosis}
In the previous section, we show a number of individual prognostic factors derived from similar patients searched through histopathological images, thus stratifying patients with different prognoses. Next, we investigate whether the integration of all identified survival-related characteristics would provide better prognostic outcomes. We establish a Lasso-Cox proportional hazards model to select the most informative factors and calculate the risk index for each patient. According to the risk index, patients were divided into low-risk or high-risk groups according to the median. Compared to the use of individual functions, the Lasso-Cox model provides significantly better patient stratification. Among the 25 prognostic factors related to survival, seven prognostic factors were selected: pathological metastasis, serum calcium, primary lymph node, platelet qualitative, cancer status, laterality, hemoglobin. Most of the pairwise mutual information values between them are smaller than those between significantly prognostic factors (Fig. 3c), which shows that the patient's multiple prognostic factors are complementary in predicting survival outcomes.

Using univariate and multivariate Cox proportional hazards analysis, we need to clarify that the Lasso-Cox risk index is independent of known prognostic factors \cite{48}(Table 2). We perform a comprehensive comparison between the lasso-Cox risk index and other known prognostic characteristics, including nine features with a low correlation coefficient(Fig. 4). Among these characteristics, the univariate Cox proportional hazard model only correlates them with survival predictions, and the multivariate Cox proportional hazard analysis shows that our Lasso-Cox is an independent prognostic factor. 

\subsection{Personalized prognostic factors weights customized for patients}
After analyzing the results of similar images, map the results to the corresponding pathological and clinical reports, and extract their prognostic factors. In the previous sections, we analysis the importance and independence of single-factor and integrating-factor prognostic factors for survival prognosis. We carry out correlation analysis on patients' prognostic factors and used regression to assign different weights to each feature.

In the results, we select three patients with different kidney cancer subtypes and grades, select a pathological image of each of them (Fig. 5), and assigned weights to their prognostic factors. We selected some positive correlation and some negative correlation. The weight with high coefficients are listed in Table 3. From our results, we can see that cancer grade, TNM staging, and cancer diagnosis time are the three most important factors, which are in line with the judgment standards of clinicians and pathologists. In addition to these three significant factors, other key diagnostic information of cancer also have different weights in different patient reports, such as whether tumor cells invade the renal vein, renal sinus, and renal capsule in renal cell carcinoma, whether the degree of  calcium ion enriches in the tumor area, whether tumor cells invade the qualitative results of lymph nodes or platelets, and hemoglobin results. These characteristics, without exception, affect the patient's survival prognosis information. From the brief weighing results we gave, we can also find that different cancer subtypes have different weight distributions. For example, the weight of tumor type of papillary cell carcinoma is higher than that of other subtypes; the tumor area of chromophobe cell carcinoma is calcified. His corresponding weight is also higher. We discussed all the weights we got with clinicians and pathologists and got their confirming that the weights of prognosis factors we get are clinically meaningful.

\section{Discussion}
During the single-feature regression analysis, we find that the following prognostic factors are ranked first: cancer neoplasm status, histological type, and grade. Using multivariate Cox regression, we found that some factors such as gender and laterality do not influence the survival outcomes, while some other factors such as the TNM stage and neoplasm histologic grade do have an influence on survival. At the overall view, the framework users shall understand that man-made factors are more highly associative because they are engineered in such ways. The gist of providing these weights gives insights into the amount of relative importance among the prognostic factors. 

Improvements to the framework could still yield a more appropriate set of patients or personalized populations with similar histology. One way is curating a larger set of data with both histopathological images with associating pathological reports and medical information from various databases. The challenging aspect of establishing a larger and more diverse data set involves different data standards and interoperability. A common medical data structure for generating pathological reports from pathological images could be developed based on this work.

It is quite clear that genomic-related diseases such as cancer could be more effectively treated with precision medicine practices. Embeded in each histopathological image, there are random visual information and extracted pathological analyses and many omics (e.g. genomics, proteomics, transcriptomics) and analytical computational biology characteristics. Integrating these “hard-to-interpret” information into our personalized pathology report could yield a more precise diagnosis and prognosis. One way to deploy the proposed technique to end-users such as pathologists oncologists is to integrate the prognostic factor weights with digital pathology reports to the existing digital pathology platform such as OpenHI. Oncologists would be able to plan more precise treatment plans according to the personalized prognostic factors with the help of the automatically predicted weights. Further work will include the fusion of omics data with images and text data for a more detailed multi-factor weight analysis of patients, which would help inexperienced pathologists or intern doctors to have a deeper understanding of pathology reports through this function. At the same time, clinicians can modify the weight of prediction in the interactive interface. The system may adaptively learn and correct itself since all predicted values will be saved to improve the prediction model and interactive interface. 

\section{Conclusion}
To our knowledge, this is the first work that jointly analyzed histopathological images with pathological reports to calculate weights of prognostic factors and predict survival outcomes based on a personalized group of histologically similar patients. The proposed framework could analyzed whole-slide images to search for similar cases. Then, the model calculated a personalized a set of weights to be attached to prognostic factors. In addition, a robust prognostic model was built to also predict personalized survival outcomes of patients using these two types of data. 

The kidney cancer data from TCGA Project was used in framework development and testing and validate the robustness of our model. In addition to renal cell carcinoma, our model can be applied to other cancers in the future. We believed that the proposed approach can be applied to other types of cancer for generating personalized prognostic factor weights. An interactive form of this framework would greatly benefit pathologists and oncologists in the future. By expanding the robustness of our model, complete and accurate pathologically assisted diagnosis can be made, effectively helping pathologists in daily tasks and specific cases.


\section*{Acknowledgment}
This work has been supported by National Natural Science Foundation of China (61772409); The consulting research project of the Chinese Academy of Engineering (The Online and Offline Mixed Educational Service System for “The Belt and Road” Training in MOOC China); Project of China Knowledge Centre for Engineering Science and Technology; The innovation team from the Ministry of Education (IRT\_17R86); and the Innovative Research Group of the National Natural Science Foundation of China (61721002). The results shown here are in whole or part based upon data generated by the TCGA Research Network: https://www.cancer.gov/tcga.



%



\end{document}